\documentclass[lettersize,journal]{IEEEtran}
\usepackage{amsmath,amsfonts}
\usepackage{algorithmic}
\usepackage{algorithm}
\usepackage{array}
\usepackage[caption=false,font=normalsize,labelfont=sf,textfont=sf]{subfig}
\usepackage{textcomp}
\usepackage{stfloats}
\usepackage{url}
\usepackage{verbatim}
\usepackage{xcolor}  
\usepackage{graphicx}
\usepackage{float}
\usepackage{cite}
\usepackage{tabu}                     
\usepackage{multirow}                 
\usepackage{multicol}                 
\usepackage{multirow}                
\usepackage{float}                    
\usepackage{makecell}                 
\usepackage{booktabs}                 

\begin{document}


\title{FedFusion: Manifold Driven Federated Learning \\ for Multi-satellite and Multi-modality Fusion} 

\author{DaiXun Li, Weiying Xie, \IEEEmembership{Senior Member,~IEEE}, Yunsong Li, \IEEEmembership{Member,~IEEE}, and Leyuan Fang, \IEEEmembership{Senior Member,~IEEE}
\thanks{This work was supported in part by the National Natural Science Foundation of China under Grant 62121001, U22B2014, and the Youth Talent Promotion Project of China Association for Science and Technology under Grant 2020QNRC001, in part by the Fundamental Research Funds for the Central Universities under Grant QTZX23048.}      
\thanks{D. Li, W. Xie, Y. Li are with the State Key Laboratory of Integrated Services Networks, Xidian University, Xi'an 710071, China (e-mail: ldx@stu.xidian.edu.cn; wyxie@xidian.edu.cn; ysli@mail.xidian.edu.cn;).}
\thanks{L. Fang is with the College of Electrical and Information Engineering, Hunan University, Changsha 410082, China (e-mail: fangleyuan@gmail.com).} }

\markboth{IEEE Transactions on Geoscience and Remote Sensing}%
{Shell \MakeLowercase{\textit{et al.}}: A Sample Article Using IEEEtran.cls for IEEE Journals}


\maketitle

\begin{abstract}
	Multi-satellite, multi-modality in-orbit fusion is a challenging task as it explores the fusion representation of complex high-dimensional data under limited computational resources. Deep neural networks can reveal the underlying distribution of multi-modal remote sensing data, but the in-orbit fusion of multimodal data is more difficult because of the limitations of different sensor imaging characteristics, especially when the multimodal data follows non-independent identically distribution (Non-IID) distributions. To address this problem while maintaining classification performance, this paper proposes a manifold-driven multi-modality fusion framework, FedFusion, which randomly samples local data on each client to jointly estimate the prominent manifold structure of shallow features of each client and explicitly compresses the feature matrices into a low-rank subspace through cascading and additive approaches, which is used as the feature input of the subsequent classifier. Considering the physical space limitations of the satellite constellation, we developed a multimodal federated learning module designed specifically for manifold data in a deep latent space. This module achieves iterative updating of the sub-network parameters of each client through global weighted averaging, constructing a framework that can represent compact representations of each client. The proposed framework surpasses existing methods in terms of performance on three multimodal datasets, achieving a classification average accuracy of 94.35$\%$ while compressing communication costs by a factor of 4.  {\color{black}Furthermore, extensive numerical evaluations of real-world satellite images were conducted on the orbiting edge computing architecture based on Jetson TX2 industrial modules, which demonstrated that FedFusion significantly reduced training time by 48.4 minutes (15.18$\%$) while optimizing accuracy.} The codes will be available
	at: \url{https://github.com/LDXDU/FedFusion}.

\end{abstract}


\begin{figure}[t]
	\centering
	\includegraphics[width=3.45in]{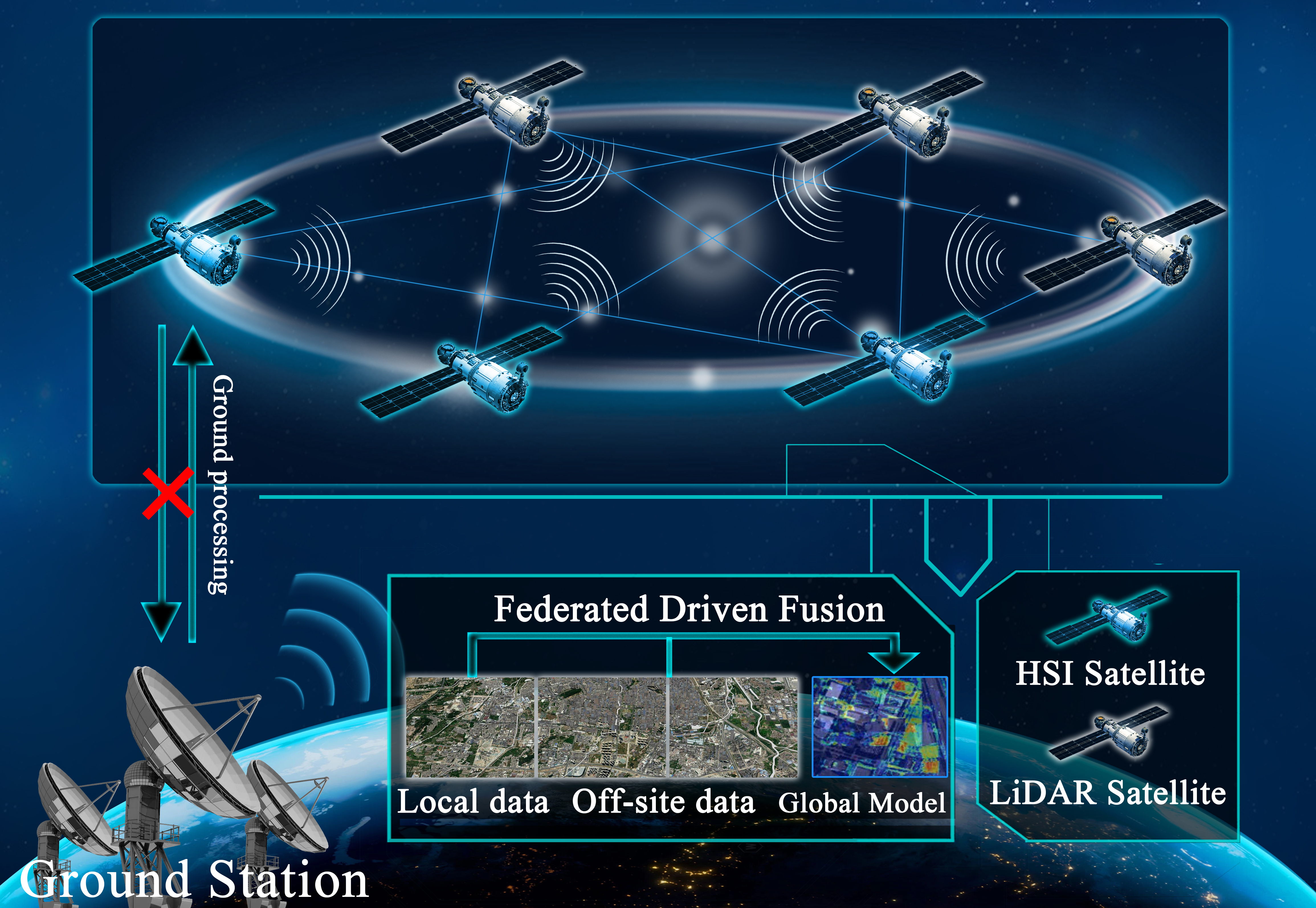}
	\caption{{\color{black}FedFusion framework at multi-satellite and multi-modality network. It includes HSI satellites and LiDAR satellites. Each client does not need to send local data back to the ground station for centralized processing, to realize the exchange of in-orbit data between multiple clients and form a global model.}}
	\label{fig-satellite}
\end{figure}

\section{Introduction}

\IEEEPARstart{W}{ith} the continuous development of low-orbit satellites, more countries and organizations are adopting multi-satellite constellations to tackle complex remote sensing tasks. These systems enable autonomous coordination of multiple remote sensing satellites\cite{5974549,6400073,9047526,7554256,9247414,zhang2023superyolo}, {\color{black}facilitating multi-modality feature classification among scientific payloads \cite{9160172,liu2022separable,liu2019unsupervised}}. Additionally, fusion of spectral and spatial information extracted from each modal data effectively enhances the accuracy of distinguishing target classes. However, some data types in multi-satellite distributed systems are sensitive in nature. For space-borne data, only level-1 or higher products are provided for public access, and level-0 data is protected as privacy by relevant laws\cite{9831991}, many jurisdictions have implemented data protection regulations (e.g., GDPR) that strictly limit the sharing of privacy-sensitive data among different clients and platforms\cite{9228139,9051171,8579471,yang2021model}. As a result, processing data closest to the source remains a challenge for remote sensing in-orbit fusion.  In addition, doing so has significant advantages for deep learning scalability and can offer benefits such as enhanced privacy, reduced latency, greater reliability, and more efficient network bandwidth utilization.

{\color{black}There are some significant challenges that need to be addressed when implementing cooperative sensing of multi-modal remote sensing constellations{\color{black}\cite{gu2021multimodal}}. Firstly, the communication window period and long-distance transmission cause limited bandwidth and weak communication between platforms, which hinders the real-time fusion of multi-modal constellations in orbit. Therefore, this leads to the current multimodal data fusion needs to be sent back to the ground for processing \cite{gu2017discriminative}, and even leads to complete interaction failure. In recent years, distributed computing has had some viable applications for all-round and high precision interaction of extensive multimodal remote sensing data. For example, Wu\emph{ et al.} \cite{7446277} proposed a hyperspectral image classification method based on cloud computing to realize the dimensionality reduction of hyperspectral images through a distributed data synchronization method. Li \emph{et al.} \cite{9438067} proposed a heterogeneous distributed fusion method to accelerate remote sensing and social media data fusion. Sun \emph{et al.} \cite{8626483} proposed a new cloud computing solution to the efficient processing to fuse the information of panchromatic images. J. M. Haut \emph{et al.} \cite{8798981} proposed a cloud-distributed model to extract the spectral channels of remote sensing datasets. All of the multi-modality fusion frameworks described above have focused on the problem of integrated data utilization. Although these distributed methods have alleviated the problem by reducing the amount of transmitted data through the use of processed features instead of raw data, the frequent transmission of data streams and the interaction patterns of high-dimensional manifold structures are still insufficient to meet the needs of multimodal fusion of remote sensing constellations in orbit. In addition, from the perspective of data privacy and communication burden, how to effectively fusion remote sensing image in-orbit has not been solved. Satellites federated learning becomes a promising solution to address these challenges as an efficient multi-clients communication scheme.}

Since Google introduced the concept of “federated learning” in 2016\cite{DBLP:journals/corr/McMahanMRA16}, federated learning has become a hot topic in many research areas, including finance and healthcare\cite{9671857,9983621,9284689,9498820,10001077,10020281}. Federated learning attempts to use technical algorithms to encrypt the constructed models. The federated parties can train model parameters without providing their original data, which is expected to produce better results for the in-orbit processing of multi-satellite distributed systems.

However, a deep learning architecture for multi-satellite multi-sensor information interaction still needs to be improved for remote sensing in-orbit satellites. It focuses on two generic questions, “How to perform multi-modality fusion private” and “How to perform multi-modality fusion efficiently”\cite{9174822}. To this end,  we systematically discuss various structures and propose network structures based on the following characteristics: (1) the transmission characteristics preserves a low-rank property lying in low-dimensional manifolds; (2)  multi-modality fusion serves in-orbit images classification and optimally updates gradient parameters. As shown in Fig. \ref{fig-satellite}, we propose a manifold-driven multimodal fusion framework for multi-satellite and multi-modality fusion, FedFusion, this method breaks the dilemma of space limitation of in-orbit multi-modality fusion. Notably, manifold driven multimodal feature fusion and low-rank subspace learning are embedded into FedFusion to estimate complex multimodal feature representations. Extensive experiments on three multimodal data demonstrate that the framework exhibits a significant advantage for generalized multi-modality learning problems. The contributions of this study can be summarized as:


\begin{itemize}
	\item[$\bullet$] We propose FedFusion, a novel manifold-driven multimodal fusion framework. It randomly samples local data and jointly estimates the salient manifold structure of each client's shallow features.   Additionally, benefited from the singular value decomposition (SVD) , it compresses the low-dimensional representation of each modal data feature matrix into a low-rank subspace through concatenative and additive methods.  This design for manifold dimensionality reduction in multimodal in-orbit fusion mitigates the disaster of dimensionality caused by the redundancy of multimodal data.
	
	\item[$\bullet$] A multimodal federated learning module has been developed specifically for deep latent space manifold data. In this module, each client executes local computations based on the weighted average of global gradients and its local modality. The updated parameters are then sent to the server for further iteration, resulting in the creation of a framework that can represent each client's compact representation. This general module can be integrated into any other multimodal fusion framework.
	
	\color{black}\item[$\bullet$] We conducted extensive experiments on real-world multimodal satellite images using a Non-IID approach on both high-performance computing platforms and orbit edge computing architectures. Our approach achieved remarkable accuracy surpassing current multimodal classification methods with minimal communication costs. Furthermore, to validate the efficiency of the federated model, we compared its training time with mainstream algorithms, resulting in optimal performance.
\end{itemize}

\section{Related Work}
\subsection{Multimodal Classification}

In the face of increasingly sophisticated and diverse application demands for remote sensing scene characterization, deep learning has made remarkable progress in addressing the technical bottlenecks that arise from the fusion processing of multimodality. In recent years, researchers have been exploring the utilization of the multimodal deep learning models to advance the classification of multimodal remote sensing images. This progress can be categorized into two primary groups.

The progress of multimodal deep learning models can be discussed from two perspectives: feature fusion methods and RS classification learning framework. Firstly, regarding feature fusion methods, traditional multi-modality classification tasks are divided into three main categories with different stages of feature fusion: (1) early fusion, (2) network layer fusion, (3) late fusion\cite{NIPS2014_7bccfde7,7989236,li2019deep}. The traditional early fusion, which integrates two modal data into a deep learning model by a direct stacking method, could have performed better in practice. Several data fusion approaches have been proposed to address this obstacle. For example, Audebert \emph{et al.} \cite{AUDEBERT201820} proposed an early fusion network based on the FuseNet principle to jointly learn more substantial multi-modality features. In addition, some methods based on multi-modality fusion between network layers have been proposed. For example, He \emph{et al.} \cite{10049798} presented a hypergraph parser to imitate guiding perception to learn intra-modal object-wise relationships. Hong \emph{et al.} \cite{9179756} proposed the classification of hyperspectral and LiDAR data through deep encoder-decoder network architecture. Late-fusion architectures are interpreted as a combination of hybrid expert and stacked generalization methods. For example, P. Bellmann \emph{et al.} \cite{9412509} proposed a sample-specific late-fusion classification architecture to improve performance accuracy. In multimodal remote sensing data fusion, network layer fusion shows better competitiveness.

Secondly, from the perspective of image classification learning framework, RS multimodal classification framework is mainly divided into traditional classifiers, classical convolutional networks and transformer networks. In terms of conventional classifiers, Guo \emph{et al.} \cite{6352721} proposed the potential of using support vector machine (SVM) and random fores to estimate forest above-ground biomass through multi-source remote sensing data. In terms of classical convolutional networks, Ding \emph{et al.} \cite{9926173} proposed a new global-local converter network for joint classification of HSI and LiDAR data, fully mining and combining the complementary information of multimodal data and local/global spectral spatial information. {\color{black}Liu  \emph{et al.} \cite{liu2022class} proposed a class-guided coupled dictionary learning method to collaboratively classify multi-spectral and hyperspectral remote sensing images, aiming to effectively leverage the combined advantages of both modalities.} In terms of transformer architecture, Roy \emph{et al.} \cite{roy2022multimodal} proposed a new multimodal fusion transformer network that uses other multimodal data as external classification markers in transformer encoders, which achieves better generalization.

In the field of RS multimodal fusion, various fusion methods can be employed including early fusion, late fusion, and network layer fusion.  However, early fusion techniques combine different modalities into a multi-channel image before network input, which can lead to issues such as the curse of dimensionality, information redundancy, and over-fitting.  Specifically, the high-dimensional input features make the model hard to train, increase computational complexity, and result in blurred object boundaries. Late fusion techniques may result in information loss, especially lower-level information. The network layer fusion employed in FedFusion can fully leverage the diversity and complementarity of different modalities to alleviate information redundancy and over-fitting issues. {\color{black} In addition},  in handling non-linear problems and learning deeper and more complex features, using  2D-CNN network architecture proves to be more effective than using traditional models such as SVM. Compared with the Transformer network architecture, a significant reduction in model parameters can be achieved while maintaining comparable accuracy.  This kind of parameter reduction improves hardware adaptability and is suitable for RS federated learning applications.

\begin{figure*}[htbp]
	\begin{center}
		\includegraphics[width=7in]{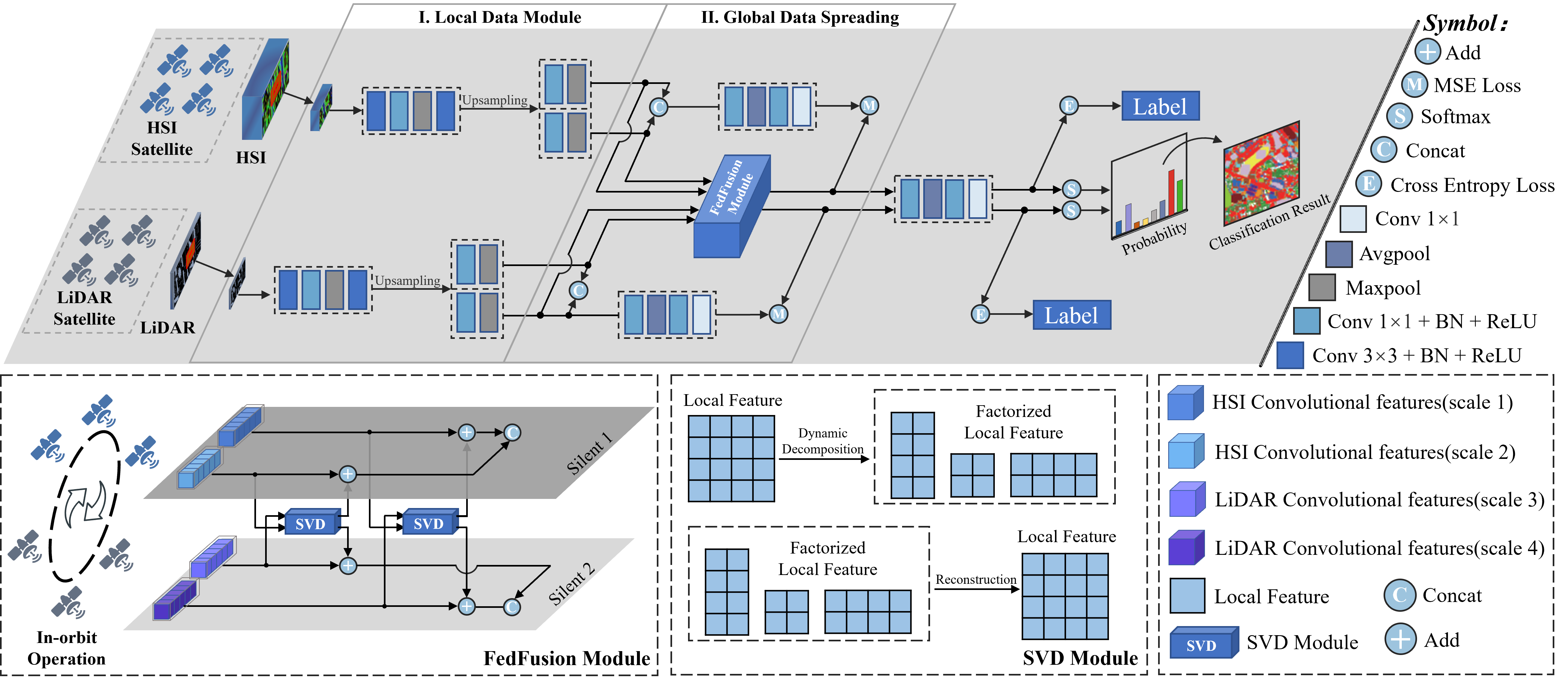}
	\end{center}
	\vspace{-0.2in}
	\caption{{\color{black}Overview of the proposed FedFusion framework. The framework includes: (1) Local Data Module, (2) Global Data Spreading, and classifier. The architecture will be optimized with respect to the mean squared error (MSE) loss of the classifier and task-specific loss for scene classification. In the global data dissemination phase, the pseudo-feature values of each client are propagated to each client through federated learning, in which singular value decomposition is used to reduce the communication load of feature value transmission.}}
	\label{fig-baseline}
	\vspace{-0.1in}
\end{figure*}

\subsection{Federated Learning}
With the growth of big data, data owners are no longer only concerned about the volume of data but are becoming more concerned about the privacy and security. Researchers have recently explored the possibility of using federated learning and developing variants to process heterogeneous data from edge devices efficiently. In the application scenario of federation learning, the differences between edge devices and collected data can make the whole training process inefficient. There are four approaches to solving the heterogeneity problem of the system: asynchronous communication, device sampling, fault-tolerance mechanisms, and model heterogeneity.

In a multi-edge device environment, asynchronous communication solutions can better address the system’s heterogeneity according to the asynchronous communication scheme. Dai \emph{et al.} \cite{dai2015high} proposed high-performance distributed machine learning at a large scale by parameter server consistency models. Also, not every device needs to be involved in every iteration of the training process. Qi \emph{et al.} \cite{qi2020privacy} proposed a recommendation algorithm that randomly selects the local gradients of clients to upload to the data center to train the global model. Similarly, we need to consider the acceptability of devices in a federated learning environment. Wang \emph{et al.} \cite{qi2020privacy} proposed a federated learning algorithm that determines the optimal trade-off between local updates and global parameter aggregation to accommodate the arithmetic limitations of edge devices. Finally, analytical data handling from different edge devices is the key to federation learning. Ren \emph{et al.} \cite{ren2020scheduling} proposed a local-global federation averaging algorithm that combines local representation learning and global model federation training to improve the model’s flexibility in handling heterogeneous data. However, these methods spend much effort extracting heterogeneous data features, and most of their architectures are based on edge-side devices. Therefore, it is significant to mine the heterogeneous feature information of remote sensing images to improve classification accuracy.

\section{Proposed Method}
\subsection{Problem Definition}

In this paper, we focus on how to train global models with different modalities of data $I_1, I_2 \in \mathbb{R}^{h \times w \times c}$ stored on heterogeneous devices to achieve classification tasks. The $h$, $w$, and $c$ denote the height, width, and the number of image channels, and $I_1(p)$ is represented as the $p$-th pixel pair of the first modality. The two modalities capture the same scene with the same label information, denoted as ${L}\in\mathbb{R}^{h \times w \times c}$ with $C\in N^*$ classification categories. The goal of image classification is to learn a model $\Theta_{\left(I_1, I_2\right)}$, and map the input images of different modalities into a new mapping $C_{\max }\left(I_1, I_2\right)$ to represent the probability of each pixel corresponding to different categories, and set the maximum probability of different categories as the threshold $\tau$ to get a binary prediction graph of hard classification, where values 1 and 0 respectively represent the category and other categories, which is defined as:
\begin{equation}
	\Theta_{\left(I_1, I_2\right)} = \begin{cases}0, & \text { if } C_{\max }\left(I_1, I_2\right)<\tau, \\ 1, & \text { otherwise. }\end{cases}
\end{equation}

Based on this basic model, we propose the FedFusion network architecture, which is shown in Fig. \ref{fig-baseline}. The designed feature fusion framework allows data fusion of multiple parties to train deep learning models without any participant revealing their raw data to the server. The dynamic feature approximation approach compresses local features so that each client’s communication reduces by up to about 4.00$\times$ per round. Therefore, the fusion future map $C_{\max }\left(I_1, I_2\right)$ in our work can be expressed as: 

\begin{equation}
	C_{\max }\left(I_1, I_2\right)=\Phi\left(I_1, I_2 \mid \alpha_1, \alpha_2\right),
\end{equation}
where a nonlinear objective model $\Phi(\cdot)$ is used to transform the image space into the classification space, and $\alpha_1, \alpha_2$ represent the corresponding parameters of two branches.

At the same time, an optimized reconstruction error loss function is proposed to optimize the proposed network.

\subsection{Manifold Driven Multi-modality Fusion Framework}
Our FedFusion framework is inspired by multi-modality cross fusion. The two branches improve model performance by sharing the output of the features from the convolutional layers, which fits right in with the privacy protection of federal learning, especially for better information fusion of heterogeneous data from different sensors.        
                                                                    
First, two sets of heterogeneous data $I_1$ and $I_2$ are put into two branches of the convolution operation to obtain the to-be-learned weights $\mathrm{w}^l$ and $b^l$. Based on the above definitions, $\mathrm{X}^l$ is the output feature map of the $l$-th convolutional layer, which is defined as:

{\color{black}
\begin{equation}
	\boldsymbol{{X}_{I_S}^l({p})={w}_{I_S}^{(l)} \cdot X^{l-1}({p})+b_{I_S}^{(l)}},
\end{equation}}

where $s= 1,2$ denotes different modalities. To accelerate the convergence of the network layer and control the gradient explosion to prevent gradient disappearance, the output results are batch normalized. It can be formulated as:
{\color{black}
\begin{equation}
	\boldsymbol{\widetilde{{X}}=}\operatorname{B N}\boldsymbol{\left({X}_{I_S}^l(p)\right)}.
\end{equation}}

Before importing $\boldsymbol{\widetilde{{X}}}$ into the next block, we have the following output after the nonlinear activation function, which is performed by ReLU, i.e.,
{\color{black}
\begin{equation}
	\boldsymbol{{X}_{I_S}^l({p})=}\operatorname{ReLU}\boldsymbol{(\widetilde{{X}})}.
\end{equation}}

At each client, the output {\color{black}$\boldsymbol{{X}_{I_S}^l({p})}$} is obtained through the convolution layer. In this work, we use up-sampling to enhance the spatial dimension of single-modal data. Specifically, we feed the output {\color{black}$\boldsymbol{{X}_{I_S}^l({p})}$} as input into two identical up-sampled module to get $\boldsymbol{{Y}^{f_{1}}}$ and $\boldsymbol{{Y}^{f_{2}}}$, then the two outputs are merged in spatial dimensions to be consistent with the dimensions of subsequent multimodal data merges.

Formally, channel upsampling can be represented as:
{\color{black}
\begin{equation}
	\boldsymbol{{Y}^{f_n}_{{I}_s}(p)=} \mathrm{UpSample}\boldsymbol{ ({X}_{I_S}^l({p})),~n=1,2,}
\end{equation}}
where {\color{black}$\mathrm{UpSample}($·$)$} is made up of two sets of {\color{black}$\mathrm{Conv_{1\times1}}($·$)$}, {\color{black}$\mathrm{BN}($·$)$}, {\color{black}$\mathrm{Relu}($·$)$} and the maximum pooled layer. The upper input needs to enter these two sets to get {\color{black}$\boldsymbol{{Y}^{f_1}_{{I}_s}(p)}$} and {\color{black}$\boldsymbol{{Y}^{f_2}_{{I}_s}(p)}$} respectively.

The fusion-layer  learns the characteristics of different modality by interactively updating the parameters of other subnetworks, the fusion output obtained can be expressed as: 

{\color{black}
\begin{equation}
	\boldsymbol{{J}_{{I_1}}}=\left[\begin{array}{ll}
		\boldsymbol{{Y}^{f_{1}}_{{I}_1}(p)} ~ + ~ \boldsymbol{{Y}^{f_{2}}_{{I}_1}(p)}
	\end{array}\right],
\end{equation}

\begin{equation}
	\boldsymbol{{J}_{{I_2}}}=\left[\begin{array}{ll}
		\boldsymbol{{Y}^{f_{1}}_{{I}_2}(p)} ~ + ~ \boldsymbol{{Y}^{f_{2}}_{{I}_2}(p)}
	\end{array}\right],
\end{equation}

\begin{equation}
	\renewcommand{\baselinestretch}{1.5}
	\boldsymbol{{J}_{\mathrm{fusion}}}=\left[\begin{array}{cc}
		\boldsymbol{{Y}^{f_{1}}_{{I}_1}(p)} ~ + ~ \boldsymbol{{Y}^{f_{2}}_{{I}_2}(p)} ,\\	[1mm]
		\boldsymbol{{Y}^{f_{1}}_{{I}_2}(p)} ~+ ~ \boldsymbol{{Y}^{f_{2}}_{{I}_1}(p)}
	\end{array}\right].
\end{equation}
}

As the multi-modality sample input from the next layer, the component of $\boldsymbol{{J}_{1}}$, $\boldsymbol{{J}_{2}}$ and $\boldsymbol{{J}_{\mathrm{fusion}}}$ shares the same training weight, and the obtained cross weight has a higher information density than the original weight.

\subsection{Multimodal Federated Learning Module}

In this section, we discuss how to share heterogeneous feature of different clients in a private way to update the model of each client.    According to the characteristics of multi-modal network-layer feature transmission of remote sensing samples, we propose a manifold driven federated learning framework. The program consists of two main components: (1) Local fusion update of data; (2) Update of the global model in each client. In each round, the server first selects a subset ${S} \subseteq[{N}]$ through uniform sampling, then sends a copy of the current server model to each of these selected clients $\boldsymbol{{y}_t}=\boldsymbol{{X}_t}$ :
\begin{equation}
	\boldsymbol{{y}_t} \leftarrow \boldsymbol{{y}_t}-\eta_i g_i\left(\boldsymbol{{y}_i}\right).
\end{equation}

In this study, we consider a multi-round federated learning scenario, where each client $I$ uses a fixed learning rate $\eta_i$ to compute the average gradient $g_i(\boldsymbol{{y}_i})$ on its local data at the current round $t$. As a typical example of vertical federated learning, remote sensing multi-modality federated learning allows servers from each client in the same scene aggregate heterogeneous data to achieve gradient and model updating. Vertical federated learning is categorized based on the characteristic dimension of the data, and relevant parts of data with identical scene but different characteristics are extracted for training. Therefore, vertical federated learning can improve the feature dimension of the original client data. In FedFusion, modal aggregation is employed to realize the privacy mechanism of each client, and train the global model by summarizing the feature information of each model. Specifically, the local server data is iterated and updated locally several times to obtain the output model $y_i$, which increases the amount of local computation. Subsequently, the local iteration of other servers is aggregated and averaged with the local gradient to update the model, thereby avoiding the transmission of original data during the training process.

\begin{algorithm}[htpb]
	\caption{Model Updating. The $s$ clients are indexed by $I$; the $B$ is the local minibatch size, the $y_0$ is the current model, and $E$ is the number of local epochs. }\label{alg:alg1}
	\begin{algorithmic}
		\STATE 
		\STATE {\textbf{Server Input:} {$\boldsymbol{y_0}$ , global step-size $\eta$, average gradient $g_i$}}
		\STATE  {\textbf{Server Executes:}}
		\STATE \hspace{0.5cm}$ \textbf{Initial $\boldsymbol{y_0}$ }  $
		\STATE \hspace{0.5cm}$ \textbf{for} $ each round ${ t}= 1, 2, ... \textbf{ do } $
		\STATE \hspace{1.0cm}$ \textbf{for} $ each client ${s \in I_s}$ in  parallel \textbf{ do } 
		\STATE \hspace{1.5cm}$\boldsymbol{{y}_{t+1}}({I}) \leftarrow$ \textbf{ClientUpdate} $\left({I}, \boldsymbol{{y}_t}\right)$
		\STATE \hspace{0.5cm}$\boldsymbol{{y}_{t+1}} \leftarrow \sum_{I=1}^S \frac{I_s}{I} \boldsymbol{{y}_{t+1}}$
    	\STATE
		\STATE \textbf{Client Update} $\left({I}, \boldsymbol{{y}}\right):$
		\STATE \hspace{0.5cm}$ \textbf{for} $ each local epoch ${ e}$ from 1 to  ${ E}$ \textbf{do } 
		\STATE \hspace{1.0cm}$ \textbf{for} $ batch ${b \in B}$ \textbf{ do } 
		\STATE \hspace{1.5cm}$\boldsymbol{{y}} \leftarrow \boldsymbol{{y}}-\eta_i g_i\left(\boldsymbol{{y}} ; b\right)$
		\STATE return $ \boldsymbol{y}$ to the server
	\end{algorithmic}
	\label{alg1}
\end{algorithm}

\begin{algorithm}[htpb]
	\caption{Gradient averaging. $s$ clients are indexed by $I$. }\label{alg:alg1}
	\begin{algorithmic}
		\STATE 
		\STATE {\textbf{Server Input:} {average gradient $g_i\left(I_S\right)$}}
		\STATE  {\textbf{Server Executes:}}
		\STATE \hspace{0.5cm}$ \textbf{for} $ each round ${ t}= 1, 2, ... \textbf{ do } $
		\STATE \hspace{1.0cm}$ \textbf{for} $ each round ${ s}= 1, 2, ... \textbf{ do } $ 
		\STATE \hspace{1.5cm}$ \textbf{if}  $ $g_i\left(I_s\right)$   is None 
		\STATE \hspace{2.0cm} Reserve  $g_i\left(I_s\right)$
		\STATE \hspace{1.5cm}$ \textbf{if}  $ $g_i\left(I_s\right)$   is not None 
		\STATE \hspace{2.0cm} $g_i\left(I_S\right) \leftarrow \frac{1}{|\mathcal{S}|} \sum_{i \in \mathcal{S}} g_i\left(I_S\right)$
	\end{algorithmic}
	\label{alg1}
\end{algorithm}

\subsection{Singular Value Decomposition via FedFusion}

In our FedFusion framework, although the volume of data features in a single mode may not be large, it incurs relatively high costs as the communication cycle increases. As a result, we aim to explore federated optimization algorithms that compress the exchanged feature map between clients to reduce computational expenses. This method gives special attention to the sparse attributes of remote sensing data, where the model parameters generated based on sparse representation exhibit low-rank properties. It identifies a subspace that captures the important structures of the feature map through random sampling at each client. The input matrix is then explicitly compressed to this low-rank subspace.  Specifically, the local feature of a single client can be divided into three smaller vectors by singular value decomposition. These vectors are then sequentially distributed to each client for reconstruction. Notably, we approximate the feature {\color{black}$\boldsymbol{{Y}^{f}_{{I}}(p)} \in \mathbb{R}^{P \times Q}$} by decomposing it into the product of three matrices:

{\color{black}
\begin{equation}
	\boldsymbol{{Y}^{f}_{{I}}(p)} \approx \boldsymbol{U_i \Sigma_i V_i},
\end{equation}
$\boldsymbol{{U}_i} \in \mathbb{R}^{P \times K}$, $\boldsymbol{\Sigma_i \in \mathbb{R}^{K \times K}}$,$\boldsymbol{\quad\boldsymbol{{V}_i} \in \mathbb{R}^{K \times Q}}$} are factorized matrices and $K$ is the number of retained singular values. In the remote sensing scenario application, the local feature of the server is decomposed in the same way and the diagonal matrix is split through the dynamic approximation method. Meanwhile, the local features are reduced by the diagonal matrix and sent to each client for recombination to realize lossless compression. The premise is that we are aware of the distribution structure of the sparse mode of remote sensing data during the design. 

\subsection{Loss Function}

The output of $\boldsymbol{{J}_{1}}$, $\boldsymbol{{J}_{2}}$ and $\boldsymbol{{J}_{\mathrm{fusion}}}$ passing through the global data spreading module are defined as $\boldsymbol{{O}_{1}}$, $\boldsymbol{{O}_{2}}$ and $\boldsymbol{{O}_{\mathrm{fusion}}}$. The overall network is trained by the following loss function:

{\color{black}
\begin{equation}
	\boldsymbol{\mathcal{L}_{\mathrm{CE}}=-\sum_{i=1}^n O_{\mathrm{t}} \log O_1},
\end{equation}

\begin{equation}
	\boldsymbol{\mathcal{L}_{\mathrm{MSE}}=\frac{1}{n} \times(  \sum(O_2 - O_1)^2+ \sum({O}_{\mathrm{fusion}} - O_1)^2)},
\end{equation}

\begin{equation}
		\boldsymbol{\mathcal{L}=\mathcal{L}_{\mathrm{CE}}+\mathcal{L}_{\mathrm{MSE}}+ \|w\|_2}.
\end{equation}
}
Here, $\boldsymbol{O_{\mathrm{t}}}$ refers to the actual value of the sample, $n$ represents the number of observed values in the dataset, and {\color{black}$\boldsymbol{\|w\|_2}$} denotes the sum of the L2 norm, which is utilized for computing the weights of all network layers. This approach is used to mitigate the issue of overfitting caused by an excessive number of model parameters.

\section{Experiments} 
\subsection{Data Description}

We are aware that modern Constellation systems and integrated unmanned aerial systems can incorporate a diverse range of HSI and LiDAR sensors, such as the Optech Titan MW (14SEN/CON340) equipped with an integrated camera and the CASI-1500 hyperspectral imager. Therefore, a detailed digital elevation model (DEM) of the surveyed region can be generated.  Furthermore, small unmanned aerial systems or satellite clusters, such as the GOMX-4B satellite constellation and DJI M600 Pro drone cluster, can accommodate HSI and LiDAR sensors, which facilitates the acquisition of two types of datasets for evaluating the fusion accuracy and joint communication performance in this study.  The following is a brief description of the data set used.

\subsubsection{The HSI-LiDAR Houston2013 dataset}The ITRES CASI-1500 imaging sensor was used to acquire the HSI product over the University of Houston campus and its surrounding rural areas in Texas, USA, which was made available for the IEEE GRSSDFC20132. The dataset comprises two data sources: a hyperspectral image with 144 bands covering a wavelength range of 364 to 1046 nm at a 10 nm spectral interval and LiDAR data with a single band. The image is sized at 349 $\times$ 1905 pixels. Additionally, the dataset investigates 15 land use land cover data (LULC) related categories in the scene. Table \ref{tab-houstondataset} displays the distributions of training and testing samples used for the classification task.

\begin{table}[H]
	\renewcommand{\arraystretch}{1.3}
	\setlength{\tabcolsep}{2.1mm}{
	\caption{A List of the Number of Training And Testing Samples for Each Class in Houston2013 Dataset}
	\label{tab-houstondataset}
	\centering
	\begin{tabular}{ccc|ccc}
		\toprule[1.2pt]
		\textbf{Land cover} & \textbf{Train} & \textbf{Test} & \textbf{Land cover} & \textbf{Train} & \textbf{Test} \\ 
		\midrule[1.2pt] 
		Background & 662013 & 652648 & Grass-healthy & 198 & 1053 \\ 
		Grass-stressed & 190 & 1064 & Grass-synthetic & 192 & 505 \\ 
		Tree & 188 & 1056 & Soil & 186 & 1056 \\ 
		Water & 182 & 143 & Residential & 196 & 1072 \\
		Commercial & 191 & 1053 & Road & 193 & 1059 \\ 
		Highway & 191 & 1036 & Railway & 181 & 1054 \\ 
		Parking-lot1 & 192 & 1041 & Parking-lot2 & 184 & 285 \\ 
		Tennis-court & 181 & 247 & Running-track & 187 & 473 \\ 
		\bottomrule[1.2pt]
	\end{tabular}}
\end{table}

\subsubsection{The Trento multimodal dataset}The Trento multimodal dataset includes hyperspectral and LiDAR data collected from Trento AISA Eagle sensors and Optech ALTM 3100EA sensors, respectively. The HSI data contains 63 bands with wavelengths ranging from 0.42 to 0.99$\mathrm{m}$, while the LiDAR data provides elevation information in a single band. The spatial resolution is 1$\mathrm{m}$ per pixel. The scene includes 6 vegetation land-cover classes that are mutually exclusive, and the pixel count is 600 × 166 pixels.

\begin{table}[htpb]
	\renewcommand{\arraystretch}{1.3}
	\setlength{\tabcolsep}{2.7mm}{
		\caption{A List of the Number of Training And Testing Samples for Each Class in Trento Dataset}
		\label{Houston}
		\centering
		\begin{tabular}{ccc|ccc}
			\toprule[1.2pt]
			\textbf{Land cover} & \textbf{Train} & \textbf{Test} & \textbf{Land cover} & \textbf{Train} & \textbf{Test} \\ 
			\midrule[1.2pt] 
			Background & 98781 & 70205 & Apples & 129 & 3905 \\ 
			Buildings & 125 & 2778 & Ground & 105 & 374 \\ 
			Woods & 188 & 1056 & Vineyard & 184 & 10317 \\
			Roads & 122 & 3052 & ~ & ~ & ~ \\
			\bottomrule[1.2pt]
	\end{tabular}}
\end{table}

\subsubsection{The MUUFL Gulfport scene dataset}The MUUFL Gulfport scene was captured over the University of Southern Mississippi campus in November 2010 by using the Reflective Optics System Imaging Spectrometer (ROSIS) sensor. The HSI of this dataset contains 325 $\times$ 220 pixels with 72 spectral bands. But the initial and final 8 bands are removed due to noise, resulting in a total of 64 bands, and the LiDAR image contains elevation data of 2 bands. The dataset includes 11 urban land-cover classes with 53687 ground truth pixels.

\begin{table}[htpb]
	\renewcommand{\arraystretch}{1.3}
	\setlength{\tabcolsep}{1.63mm}{
		\caption{A List of the Number of Training And Testing Samples for Each Class in the MUUFL Dataset}
		\label{Houston}
		\centering
		\begin{tabular}{ccc|ccc}
			\toprule[1.2pt]
			\textbf{Land cover} & \textbf{Train} & \textbf{Test} & \textbf{Land cover} & \textbf{Train} & \textbf{Test} \\ 
			\midrule[1.2pt] 
			Background & 68817 & 20496 & Trees & 1162 & 22084  \\ 
			Grass-Pure & 214 & 4056 & Grass-Groundsurface & 344 & 6538  \\ 
			Dirt-And-Sand & 91 & 1735 & Road-Materials & 334 & 6353  \\ 
			Water & 23 & 443 & Buildings'-Shadow & 112 & 2121  \\ 
			Buildings & 312 & 5928 & Sidewalk & 69 & 1316  \\ 
			Yellow-Curb & 9 & 174 & ClothPanels & 13 & 256  \\ 
			\bottomrule[1.2pt]
	\end{tabular}}
\end{table}

{\color{black} \subsubsection{The Augsburg on-site multimodal satellite dataset}To simulate real in-orbit federated application scenarios similar to FedFusion, we employed an on-site multi-modal satellite dataset to validate the framework's effectiveness. This dataset comprises synthetic aperture radar (SAR) data and HSI data captured over the Augsburg city in Germany. For the SAR data, we obtained the Ground Range Detected (GRD) product in Interferometric Wide (IW) swath mode using the Sentinel-1 satellite platform. This product includes dual-polarization SAR data with VV and VH channels. It is important to note that no preprocessing operations were performed on the data that would affect the nature of the on-site data, except for precise orbit profiling, radiometric calibration, and terrain correction. Additionally, we collected the HSI images dataset using the Sentinel-2 satellite platform in near real-time. The images represent bottom-of-atmosphere (BOA) reflectance with 12 spectral bands ranging from 440 to 2200 nanometers.

\begin{table}[htpb]
	\color{black}
	\renewcommand{\arraystretch}{1.3}
	\setlength{\tabcolsep}{1.63mm}{
		\caption{A List of the Number of Training And Testing Samples for Each Class in On-site Augsburg Dataset}
		\label{Houston}
		\centering
		\begin{tabular}{ccc|ccc}
			\toprule[1.2pt]
			\textbf{Land cover} & \textbf{Train} & \textbf{Test} & \textbf{Land cover} & \textbf{Train} & \textbf{Test} \\ 
			\midrule[1.2pt] 
			Background & 68816 & 20497 & Forest & 1162 & 22084  \\ 
			Residential-Area & 344 & 6538 & Industrial-Area & 91 & 1735  \\ 
			Low-Plants & 334 & 635 & Allotment & 23 & 443  \\ 
			Commercial-Area & 112 & 2121 & Water & 312 & 5928  \\ 
			\bottomrule[1.2pt]
	\end{tabular}}
\end{table}

 }

\begin{figure*}[t]
	\centering
	\includegraphics[scale=1.076]{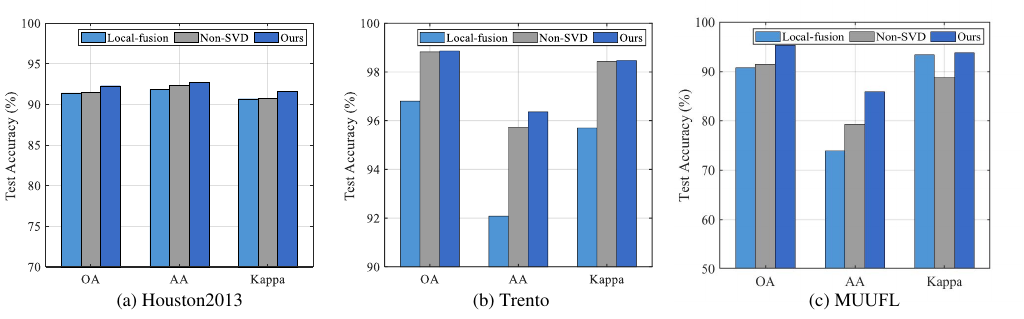}
	\centering
	\caption{{\color{black}Overview of the ablation effect of FedFusion framework and SVD module on the three datasets.}}
	\label{fig-a1}
\end{figure*}

\subsection{Evaluation Metrics and Parameter Setting}

\subsubsection{Evaluation Metrics} In terms of testing image classification performance, three commonly used indexes: Overall Accuracy (OA), Average Accuracy (AA), Kappa ($\kappa$) and Class Accuracy(CA) are calculated to quantify classification performance. They can be formulated by the following equations. CA represents the model's ability to classify and recognize each category in the data set.

\begin{equation}
	\begin{gathered}
		O A=\frac{N_c}{N_a}, \\
	\end{gathered}
\end{equation}

\begin{equation}
	\begin{gathered}
		A A=\frac{1}{C} \sum_{i=1}^c \frac{N_c^i}{N_a^i}, \\
	\end{gathered}
\end{equation}
\begin{equation}
	\begin{gathered}
		\kappa=\frac{O A-P_e}{1-P_e}, \\
	\end{gathered}
\end{equation}

and

\begin{equation}
	\begin{gathered}
		P_e=\frac{N_r^1 \times N_p^1+\cdots N_r^i \times N_p^i+\cdots+N_r^C \times N_p^C}{N_a \times N_a}.
	\end{gathered}
\end{equation}

$N_c$ and $N_a$ represent the number of correctly classified samples and the total number of samples, respectively, while $N_c^i$ and $N_a^i$ correspond to $N_c$ and $N_a$ of each category, respectively. $P_e$ in $\kappa$ is defined as the assumed probability of contingency coincidence. 

At the same time, by calculating the traffic transmitted in the training stage, namely the modal characteristic parameter values of other clients,  the communication ratio between each client and the original model is obtained to verify the effectiveness of the communication lightweight module.

\subsubsection{Parameter Setting} {\color{black}Our framework FedFusion is tested in two scenarios. In the toy experiments, we select 16 clients, each of which is equipped with NVIDIA A100 Tensor Core GPU and uses Pytorch platform, using NCCL as communication backend. These clients simulated eight hyperspectral imaging satellites and eight LiDAR imaging satellites and were randomly divided into a training set containing 95$\%$ data and a validation set containing 5$\%$ data. 
	
In the simulated on-orbit experiment, according to the latest research on the construction of satellite orbit hardware system\cite{so2022fedspace,denby2020orbital}, we build a federated satellite constellation system that is close to the resources of satellites in orbit.
In the construction of computer hardware, we are consistent with the existing satellite on-board computing resources. In this work, we characterize on-board computing with a Jetson TX2 industrial module. Jetson TX2 is a mobile GPU containing high performance, low power, and efficiency, and its industrial variant is designed for extreme temperature environments. Its small size allows the integration of all other necessary components within a 1U volume, and the 3U cube contains a 2U camera system. The 7.5W power pattern closely matches the 7.1W input power provided by the surface-mounted solar panels. The OEC computing model supports computing systems other than Jetson by varying the input performance and energy parameters.
On satellite data acquisition, we use the Sentinel-1 and the Sentinel-2 satellite real-time platform to evaluate the federated space. Specifically, we consider a constellation of 8 satellites, each satellite collects and stores its proprietary remote sensing data. These satellites acquired SAR images and HSI for the city of Augsburg via Sentinel-1 and Sentinel-2 satellite real-time platforms, respectively, and distributed them to 4 analog satellite industrial modules. We will use this to simulate four HSI satellites and four LiDAR satellites and run our experiments with Non-IID data distributions. Nevertheless, transmitting all satellite data to ground stations is often deemed impractical or unfeasible. To address this challenge, we leverage federated learning (FL), where a server coordinates the training process with a set of clients without sharing the clients' training data. We let all the simulated satellites act as a single FL client, and each server achieves a fused direct connection, because it is much faster than the satellite back to the ground and back to the satellite. Each FL client manages the learning process and maintains the current version of the global model and sends it to the other satellites, it also receives model updates and aggregates the updates into the global model. At the same time, we take into account the heterogeneous connections between satellites. According to the satellite connection histogram of FedSpace, the number of connections of each satellite may be quite different due to the trajectory window period. Therefore, in each round of communication fusion, the server first selects a set of working satellite clients by uniform sampling. A copy of the current server model is then sent to each of these selected clients for fusion. This heterogeneous connectivity allows updating the global model without waiting for other satellites to introduce serious obsolescence.
}

 In the training phase, the learning strategy of Adam optimizer with second-order momentum was used, and the learning rate is updated by StepLR strategy. That means the learning rate is adjusted by a fixed interval, with an interval of 60 epochs in 300 epochs totally. The initial learning rate and gamma are set to 0.001 and 0.5, respectively. The training batch of the dataset are set to 64, and the momentum parameter was set to 0.9.  Meanwhile, L2 norm regularization is used to prevent model overfitting. It is worth noting that due to the larger size and increased number of categories in the MUUFL dataset compared to others, it has fewer labels available. Therefore, we adjusted the batch size to 128 and optimized the ReduceLROnPlateau learning rate adjustment strategy.

\subsection{Ablation Study}
Firstly, the effectiveness of the proposed method is verified by designing a series of ablation experiments on three datasets.

\subsubsection{\textbf{Validation of the Baseline Framework}}{\color{black}In Fig. \ref{fig-a1}, the reasoning ability of different basic frameworks, namely local feature fusion framework (Local-fusion), federated feature fusion framework (Non-SVD), and our FedFusion feature fusion framework, is evaluated from three indicators: OA, AA, and $\kappa$. Local-fusion is a feature-level fusion model that operates between layers within a single client, using a 2D-CNN architecture. In contrast, Non-SVD is a non-lightweight federated feature fusion model. In the three datasets, while the feature transmission fusion network showed significant improvement compared to the traditional local feature fusion network, the FedFusion network further increased the accuracy compared to the federal framework. Specifically, in regard to the Houston2013 dataset, our method demonstrates a 0.87$\%$ increase in OA relative to Local-fusion and an 0.84$\%$ improvement compared to Non-SVD. Correspondingly, for the Trento dataset, our method shows a 2.06$\%$ and 0.04$\%$ increase in OA compared to Local-fusion and Non-SVD, respectively. Additionally, the MUUFL dataset highlights an 1.11$\%$ and 0.33$\%$ rise in OA for FedFusion compared to Local-fusion and Non-SVD, respectively. These outcomes validate the use of FedFusion as a fundamental framework for federated learning. In addition, the classification performance by federated learning is generally better than that based on single client communication, and our method brings incremental results to OA, AA and $\kappa$.}

\begin{table}[h]
	\small
	\renewcommand{\arraystretch}{1.2}
	\centering
	\setlength{\tabcolsep}{1.6mm}{
		\caption{Performance (\%) of Removing Single Modality in Three Datasets}
		\label{tab-a2}
		\begin{tabular}{c|cc|cc|cc}
			\toprule[1.2pt]
			\textbf{Data} & \multicolumn{2}{c|}{Houston2013} & \multicolumn{2}{c|}{Trento} & \multicolumn{2}{c}{MUUFL} \\
			\cmidrule(lr){2-3}\cmidrule(lr){4-5}\cmidrule(lr){6-7}
			\textbf{No.}            & HSI            ~& LiDAR          ~& HSI          ~& LiDAR      ~& HSI         ~& LiDAR       \\
			\midrule[1.2pt] 
			1              & 82.62          & 45.87          & 89.72        & 98.82       & 95.56       & 86.22       \\
			2              & 85.06          & 47.91          & 89.20        & 94.82       & 73.94       & 48.10        \\
			3              & 95.45          & 41.58          & 95.72        & 66.04       & 79.43       & 62.86       \\
			4              & 92.71          & 76.14          & 82.28        & 77.62       & 81.56       & 0.00           \\
			5              & 99.62          & 59.75          & 95.42        & 86.89       & 88.98       & 90.54       \\
			6              & 95.10          & 79.72          & 71.56        & 73.53       & 0.00           & 39.05       \\
			7              & 83.77          & 72.01          & ---        &  ---           & 75.53       & 26.54       \\
			8              & 74.93          & 81.20          & ---          & ---            & 85.58       & 67.68       \\
			9              & 77.53          & 51.84          &  ---            &   ---          & 56.69       & 15.12       \\
			10             & 63.32          & 64.38          &   ---           &   ---          & 0.00           & 0.00         \\
			11             & 71.44          & 74.10          &   ---           &  ---           & 74.61       & 0.00           \\
			12             & 85.40          & 39.58          &   ---           &  ---           &  ---           &  ---           \\
			13             & 92.98          & 70.53          &   ---           &  ---           &  ---           &  ---           \\
			14             & 100.00            & 53.85          &   ---           &  ---           &   ---          &  ---           \\
			15             & 99.37          & 93.23          &    ---          &  ---           &  ---           &  ---           \\
			\midrule[1.2pt] 
			OA(\%)         & 83.73          & 62.01          & 87.59        & 84.74       & 86.22       & 70.19       \\
			AA(\%)         & 86.62          & 63.45          & 87.32        & 82.95       & 64.72       & 39.65       \\
			$\kappa$(\%)      & 82.41          & 58.90           & 83.68        & 80.24       & 81.69       & 59.84  	  \\
			\bottomrule[1.2pt] 
	\end{tabular}}
	\vspace{-0.1in}
\end{table}

\subsubsection{\textbf{Impact of Removing Single Modality}}Table \ref{tab-a2} illustrates the impact of removing single modality, highlighting the effectiveness of multi-modal feature-level fusion across all three datasets. For the three datasets, the accuracy of OA, AA and $\kappa$ of HSI are 85.75$\%$, 78.55$\%$ and 82.59$\%$, respectively. Meanwhile, for LiDAR, the average accuracy of OA is 72.31$\%$, AA is 62.02$\%$, and $\kappa$ is 66.33$\%$. For the three datasets, the accuracy of FedFusion’s OA, AA and $\kappa$ are 94.28$\%$, 88.57$\%$ and 93.02$\%$, respectively. These results suggest that multi-modal feature fusion significantly improves accuracy and supports its importance in remote sensing tasks.

\begin{figure*}[htpb]
	\centering
	\includegraphics[scale=1]{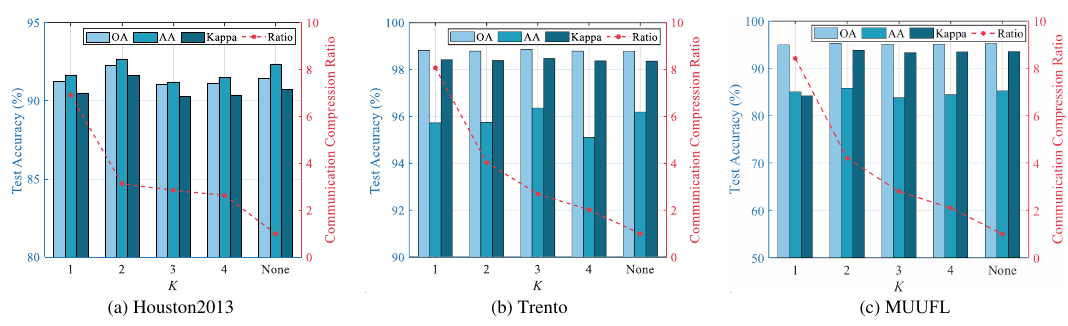}
	\centering
	\vspace{-0.2in}
	\caption{Overview of the $K$ on the singular value decomposition. The horizontal axis represents scenarios with and without SVD ($K=1, 2, 3, 4$), while the precision effect of lightweight transmission is measured using OA, AA, and $\kappa$.}
	\vspace{-0.1in}
	\label{fig-a31}
\end{figure*}

\subsubsection{\textbf{Impact of $K$ on the Dynamic Feature Approximation}}Fig. \ref{fig-a31} illustrates the impact of Manifold Driven Dynamic Feature Approximation on the accuracy of scene classification in three datasets, including non-use the method. In our approach, given an input patch size of 7$\times$7, SVD's $K$ values range from 1 to 4 based on two 2$\times$2 subsampling and maximum pooling. The line graph demonstrates that the removal of redundant feature map information via SVD significantly enhances classification accuracy while minimizing client communication. In the Houston2013 dataset, our approach yields the best accuracy results with $K=2$, leading to a 0.84$\%$ increase in OA, 0.32$\%$ increase in AA, and a 0.92$\%$ increase in $\kappa$ than without using dynamic feature approximation.


{\color{black}Moreover, Fig. \ref{fig-a31} shows the ratio of original communication transmission volume to that of lightweight and compressed traffic at $K=1,2,3,4$. It is essential to note that the Float16 data type is used in federated transmission of feature maps between clients. The experimental results show that before using SVD, the communication cost on the Houston2013 dataset, Trento dataset, and MUUFL dataset was 45.97MB, 59.58MB, and 110.26MB, respectively. When $K=1,2,3,4$, the average communication compression ratios for the three datasets are 7.801, 3.792, 2.786, and 2.252, respectively. Combining the results of Fig. \ref{fig-a31}, it is evident that a $K$ value of 2 reaches the optimal balance between client communication transmission volume and classification accuracy for all three datasets. To be specific, the communication cost of these three datasets is reduced to 14.68MB, 14.75MB, and 26.21MB, respectively. While maintaining the best accuracy, the communication cost is reduced by more than three times, greatly improving FedFusion's processing efficiency.}

\begin{table*}[htbp]
	\small
	\renewcommand{\arraystretch}{1.2}
	\centering
	\setlength{\tabcolsep}{4.35mm}{
		\caption{OA, AA and $\kappa$ Values on the Houston2013 Dataset (in $\%$) by Considering HSI and LiDAR Data}
		\label{tab-vsh}	
		\begin{tabular}{c|c|c|c|c|c|c|c|c|c}
			\toprule[1.2pt]
			\textbf{No.} & \textbf{Class} & \textbf{SVM} & \textbf{CNN-2D} & \textbf{RNN} & \textbf{Cross} & \textbf{CALC} & \textbf{ViT} & \textbf{MFT} & \textbf{FedFusion} \\ 
			\cmidrule(lr){1-10}
			1 & Healthy grass & 79.87 & \textbf{88.86} & 79.23 & 80.06 & 78.63 & 82.59 & 82.34 & 81.58 \\ 
			2 & Stressed grass  & 79.14 & 86.18 & 81.42 & \textbf{100.00} & 83.83 & 82.33 & 88.78 & 96.80 \\ 
			3 & Synthetis grass & 61.25 & 45.35 & 38.75 & \textbf{98.22} & 93.86 & 97.43 & 98.15 & 97.23 \\ 
			4 & Tree & 85.23 & 75.47 & 88.35 & 95.83 & 86.55 & 92.93 & 94.35 & \textbf{98.20} \\ 
			5 & Soil & 87.22 & 87.12 & 90.91 & \textbf{99.43} & 99.72 & 99.84 & 99.12 & 98.48 \\ 
			6 & Water & 61.54 & 72.03 & 77.39 & 94.41 & 97.90 & 84.15 & \textbf{99.30} & 95.80 \\ 
			7 & Residential & 81.53 & 68.13 & 65.76 & 95.15 & 91.42 & 87.84 & 88.56 & \textbf{95.34} \\ 
			8 & Commercial & 18.80 & 23.36 & 35.61 & 92.02 & \textbf{92.88} & 79.93 & 86.89 & 86.99 \\ 
			9 & Road & 68.27 & 59.05 & 72.05 & \textbf{98.68} & 87.54 & 82.94 & 87.91 & 91.60 \\ 
			10 & Highway & 48.46 & 32.37 & 34.04 & 60.71 & 68.53 & 52.93 & 64.70 & \textbf{88.09} \\ 
			11 & Railway & 38.71 & 43.33 & 36.78 & 88.71 & 93.36 & 80.99 & \textbf{98.64} & 89.56 \\ 
			12 & Park lot 1 & 57.54 & 35.57 & 58.44 & 81.94 & \textbf{95.10} & 91.07 & 94.24 & 94.52 \\ 
			13 & Park lot 2 & 65.96 & 52.28 & 52.61 & 91.23 & 92.98 & 87.84 & 90.29 & \textbf{93.33} \\ 
			14 & Tennis court & 83.40 & 70.58 & 81.24 & 97.57 & \textbf{100.00} & \textbf{100.00} & 99.73 & 97.17 \\ 
			15 & Running track & 48.41 & 45.24 & 40.94 & \textbf{98.73} & 99.37 & 99.65 & 99.58 & 84.78 \\ 
			\midrule[1.2pt]
			~ & OA(\%) & 64.18 & 59.06 & 62.61 & 90.34 & 88.97 & 85.05 & 89.80 & \textbf{92.30} \\ 
			~ & AA(\%) & 64.36 & 58.99 & 62.24 & 91.51 & 90.78 & 86.83 & 91.54 & \textbf{92.69} \\ 
			~ & $\kappa$(\%) & 61.30 & 55.70 & 59.64 & 89.53 & 88.06 & 83.84 & 88.93 & \textbf{91.65} \\ 
			\bottomrule[1.2pt]
	\end{tabular}}
\end{table*}

\begin{figure*}[htbp]
	\centering
	\includegraphics[scale=0.58]{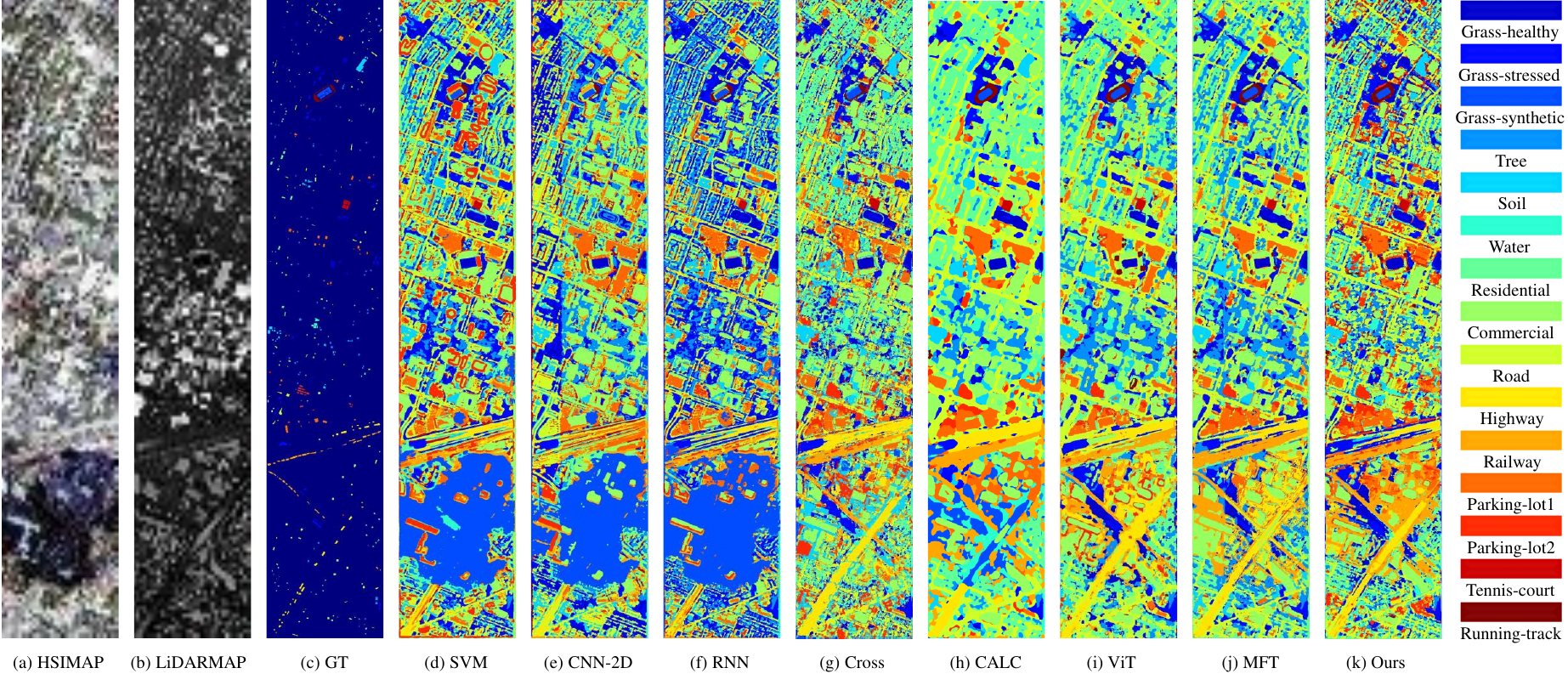}
	\centering
	\vspace{-0.2in}
	\caption{Visualization of false-color HSI and LiDAR images using different comparison methods based on Houston2013 dataset.}
	\vspace{-0.1in}
	\label{vis-h}
\end{figure*}

\subsection{Comparisons with Previous Methods}

The accuracy performance of the proposed model as well as other models on the Houston2013, Trento, and MUUFL datasets is presented in Tables \ref{tab-vsh}, \ref{tab-vst}, and \ref{tab-vsm} respectively. The best results are denoted in bold. In selecting comparison methods, we consider classic deep learning techniques such as SVM\cite{melgani2004classification}, CNN-2D\cite{makantasis2015deep}, and RNN \cite{cho2014properties}, as well as mainstream remote sensing multi-modal methods including Cross\cite{hong2020more}, CALC\cite{9926173}, and transformer methods such as ViT\cite{dosovitskiy2020image} and MFT\cite{roy2022multimodal}. Our evaluation demonstrates that the proposed method not only achieves 16 clients federated lightweight communication, but also outperforms other methods by achieving the highest OA, AA, and $\kappa$ scores across most classification tasks.

Overall, FedFusion achieves a well-balanced trade-off between communication efficiency and classification accuracy, outperforming other existing networks. The proposed model demonstrated dominant performance in all-classes accuracy on the Houston2013 dataset, surpassing mainstream methods in terms of OA, AA, and $\kappa$ coefficient. Although traditional methods exhibit higher accuracy when incorporating HSI and LiDAR data, FedFusion still shows significant improvement compared to other models. FedFusion outperforms the best algorithm MFT in transformer, with a substantial improvement of 2.50$\%$ in OA, 1.15$\%$ in AA, and 2.72$\%$ in $\kappa$ coefficient. Additionally, FedFusion's performance is slightly superior to the best traditional algorithm (OA 0.46$\%$, 1.18$\%$ AA, 2.12$\%$ $\kappa$ coefficient). {\color{black}Table \ref{tab-vst} demonstrates that FedFusion exhibits higher OA and AA than other classifiers on the Trento dataset, with marginally lower AA. The experimental results of the MUFFL dataset are shown in Table \ref{tab-vsm}. The OA, AA, and $\kappa$ of FedFusion reach 95.27$\%$, 85.82$\%$, and 93.75$\%$, respectively, which is better than the best MFT method at present. However, the transformer network (MFT) outperforms FedFusion in some aspects of performance,} which might be due to its larger model structure by approximately ten times the parameter size of the FedFusion model. However, this contradicts the intended use case for federated learning, which is suitable for in-orbit learning and processing of satellite clusters in remote sensing applications. Therefore, our FedFusion model is more suitable for practical remote sensing scenarios than other models.

{\color{black}Meanwhile, the proposed FedFusion framework is designed for the fusion of inertial multimodal on-site data. To verify the effectiveness of the framework on in-orbit data in the field, we conducted validation experiments on the on-site multimodal satellite dataset in Augsburg and compared it with other existing multimodal fusion methods. As shown in Table \ref{onsite}, the performance accuracy of FedFusion on the in-orbit satellite dataset is superior to other fusion methods, with an OA of 92.46$\%$, which outperforms the CALC method by 0.73$\%$ and the MFT method by 1.67$\%$. Moreover, FedFusion achieves comprehensive performance improvement in terms of AA, surpassing the best-performing methods by 4.81$\%$.}

\begin{table*}[htbp]
	\small
	\renewcommand{\arraystretch}{1.2}
	\centering
	\setlength{\tabcolsep}{4.5mm}{
		\caption{OA, AA and $\kappa$ Values on the Trento dataset (in $\%$) by Considering HSI and LiDAR Data}
		\label{tab-vst}	
		\begin{tabular}{c|c|c|c|c|c|c|c|c|c}
			\toprule[1.2pt]
			\textbf{No.} & \textbf{Class} & \textbf{SVM} & \textbf{CNN-2D} & \textbf{RNN} & \textbf{Cross} & \textbf{CALC} & \textbf{ViT} & \textbf{MFT} & \textbf{FedFusion} \\ 
			\cmidrule(lr){1-10}
			1 & Apple trees & 97.44 & 96.98 & 91.75 & 99.87 & 98.62 & 90.87 & 98.23 & \textbf{99.56} \\ 
			2 & Buildings & 98.12 & 97.56 & \textbf{99.47} & 98.63 & 99.96 & 99.32 & 99.34 & 98.34 \\ 
			3 & Ground & 56.15 & 55.35 & 79.23 & \textbf{98.93} & 72.99 & 92.69 & 89.84 & 77.81 \\
			4 & Woods & 97.53 & 99.66 & 99.58 & 99.22 & \textbf{100.00} & \textbf{100.00} & 99.82 & 99.21 \\ 
			5 & Vineyard & 98.13 & 99.56 & 98.39 & 98.05 & 99.44 & 97.77 & 99.93 & \textbf{99.88} \\ 
			6 & Roads & 78.96 & 76.91 & 85.86 & 89.42 & 88.76 & 86.72 & 88.72 & \textbf{92.69} \\ 
			\midrule[1.2pt]
			~ & OA(\%) & 95.33 & 96.14 & 96.43 & 97.82 & 98.11 & 96.47 & 98.32 & \textbf{98.65} \\ 
			~ & AA(\%) & 87.72 & 87.67 & 92.38 & 97.35 & 93.30 & 94.56 & \textbf{95.98} & 94.58 \\ 
			~ & $\kappa$(\%) & 93.76 & 94.83 & 95.21 & 97.09 & 97.46 & 95.28 & 97.75 & \textbf{98.19} \\ 
			\bottomrule[1.2pt]
	\end{tabular}}
\end{table*}

\begin{figure*}[htbp]
	\centering
	\includegraphics[scale=0.58]{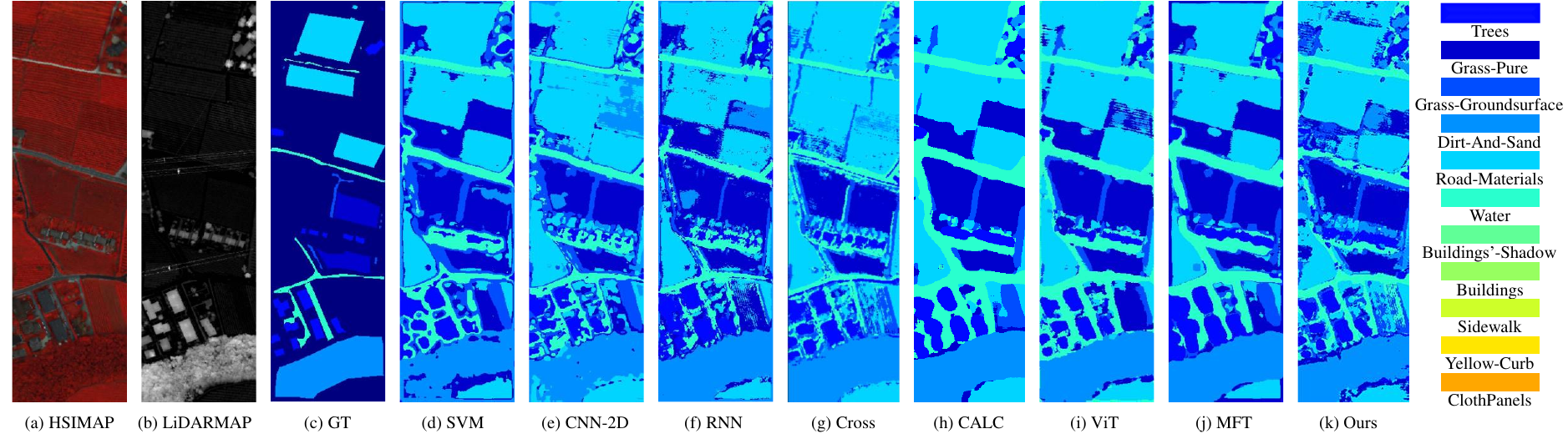}
	\centering
	\vspace{-0.2in}
	\caption{Visualization of false-color HSI and LiDAR images using different comparison methods based on Trento dataset.}
	\vspace{-0.1in}
	\label{vis-t}
\end{figure*}

\begin{table*}[htpb]
	\small
	\renewcommand{\arraystretch}{1.2}
	\centering
	\setlength{\tabcolsep}{4.1mm}{
		\caption{OA, AA and $\kappa$ Values on the MUUFL Dataset (in $\%$) by Considering HSI and LiDAR Data}
		\label{tab-vsm}	
		\begin{tabular}{c|c|c|c|c|c|c|c|c|c}
			\toprule[1.2pt]
			\textbf{No.} & \textbf{Class} & \textbf{SVM} & \textbf{CNN-2D} & \textbf{RNN} & \textbf{Cross} & \textbf{CALC} & \textbf{ViT} & \textbf{MFT} & \textbf{FedFusion} \\ 
			\cmidrule(lr){1-10}
			1 & Trees & 96.63 & 95.79 & 95.84 & \textbf{98.55} & 97.31 & 97.85 & 97.90 & {\color{black}98.46} \\ 
			2 & Mostly grass & 59.25 & 72.76 & \textbf{81.93} & 65.85 & 93.00 & 76.06 & 92.11 & {\color{black}91.03} \\
			3 & Mixed ground surface & 81.46 & 78.92 & 80.47 & 79.24 & 91.57 & 87.58 & 91.80 & {\color{black}\textbf{92.26}} \\ 
			4 & Dirt and sand & 73.54 & 83.59 & 87.01 & 69.22 & 95.10 & 92.05 & 91.59 & {\color{black}\textbf{95.33}} \\ 
			5 & Road & 83.79 & 78.29 & 90.65 & 97.12 & \textbf{95.91} & 94.71 & 95.60 & {\color{black}95.36} \\  
			6 & Water & 15.35 & 50.34 & 54.25 & 60.95 & \textbf{99.32} & 82.02 & 88.19 & {\color{black}95.94} \\  
			7 & Building shadow & 77.04 & 79.70 & 81.24 & 64.55 & \textbf{92.69} & 87.11 & 90.27 & {\color{black}90.76} \\ 
			8 & Building & 86.94 & 71.95 & 88.39 & 90.30 & 98.45 & 97.60 & 97.26 & {\color{black}\textbf{98.58}} \\  
			9 & Sidewalk & 21.28 & 43.92 & 60.54 & 32.75 & 51.60 & 57.83 & 61.35 & {\color{black}\textbf{71.73}} \\  
			10 & Yellow curb & 0.00 & 12.45 & 26.44 & 0.00 & 0.00 & \textbf{31.99} & 17.43 & {\color{black}24.71} \\ 
			11 & Cloth panels & 62.89 & 26.82 & 87.50 & 43.36 & 0.00 & 58.72 & 72.79 & {\color{black}\textbf{89.84}} \\  
			\midrule[1.2pt]
			~ & OA(\%) & 84.24 & 83.40 & 88.79 & 87.29 & 93.94 & 92.15 & 94.34 & {\color{black}\textbf{95.27}} \\ 
			~ & AA(\%) & 59.83 & 63.14 & 75.84 & 63.81 & 74.09 & 78.50 & 81.48 & {\color{black}\textbf{85.82}} \\ 
			~ & $\kappa$(\%) & 78.80 & 77.94 & 85.18 & 82.75 & 92.00 & 89.56 & 92.51 & {\color{black}\textbf{93.75}} \\
			\bottomrule[1.2pt]
	\end{tabular}}
\end{table*}

\begin{table*}[htpb]
	\color{black}
	\small
	\renewcommand{\arraystretch}{1.2}
	\centering
	\setlength{\tabcolsep}{4.1mm}{
		\caption{OA, AA and $\kappa$ Values on the On-site Augsburg Dataset (in $\%$) by Considering HSI and SAR Data}
		\label{onsite}	
		\begin{tabular}{c|c|c|c|c|c|c|c|c|c}
			\toprule[1.2pt]
			\textbf{No.} & \textbf{Class} & \textbf{SVM} & \textbf{CNN-2D} & \textbf{RNN} & \textbf{Cross} & \textbf{CALC} & \textbf{ViT} & \textbf{MFT} & \textbf{FedFusion} \\ 
			\cmidrule(lr){1-10}
			1 & Forest & 91.51 & 90.96 & 87.60 & 94.24 & 94.34 & 90.01 & 94.65 & \textbf{97.85}  \\ 
			2 & Residential-Area & 86.32 & 82.07 & 95.26 & 97.41 & \textbf{98.24} & 94.27 & 96.90 & 97.15  \\ 
			3 & Industrial-Area & 10.03 & 48.31 & 51.35 & 75.98 & \textbf{78.07} & 64.62 & 69.80 & 76.63  \\
			4 & Low-Plants & 62.78 & 61.05 & 91.14 & 89.81 & 94.57 & 85.53 & 93.98 & \textbf{97.42}  \\ 
			5 & Allotment & 22.75 & 18.48 & 10.52 & 42.26 & 28.68 & 29.00 & 32.70 & \textbf{61.22}  \\ 
			6 & Commercial-Area & 0.00 & 12.11 & 0.00 & 8.00 & 2.20 & 20.41 & 10.52 & \textbf{34.18}  \\
			7 & Water & 11.28 & 29.13 & 24.33 & 17.05 & 42.27 & 44.17 & 23.98 & \textbf{64.69}  \\ 
			\midrule[1.2pt]
			~ & OA(\%) & 71.65 & 71.79 & 83.42 & 89.37 & 91.73 & 86.10 & 90.49 & \textbf{92.46}  \\ 
			~ & AA(\%) & 40.67 & 48.88 & 40.75 & 60.68 & 62.62 & 61.14 & 60.36 & \textbf{67.43}  \\ 
			~ & $\kappa$(\%) & 58.62 & 57.84 & 75.26 & 84.77 & 87.98 & 80.06 & 86.26 & \textbf{89.14}  \\ 
			\bottomrule[1.2pt]
	\end{tabular}}
\end{table*}

\begin{figure*}[htbp]
	\centering
	\includegraphics[scale=0.58]{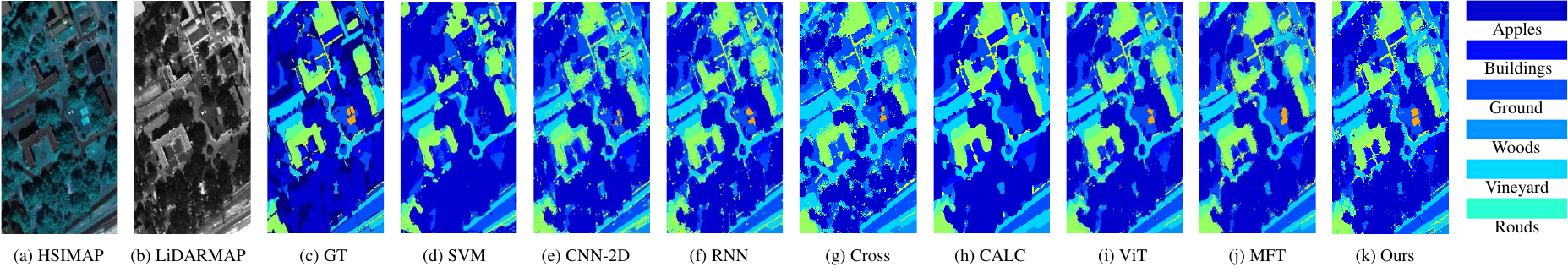}
	\centering
	\vspace{-0.2in}
	\caption{Visualization of false-color HSI and LiDAR images using different comparison methods based on the MUUFL dataset.}
	\vspace{-0.1in}
	\label{vis-m}
\end{figure*}

\begin{table}[h]
	\color{black}
	\small
	\renewcommand{\arraystretch}{1.2}
	\caption{Performance of Each Method on Federated and Local Algorithms}
	\centering
	\label{speed}
	\begin{tabular}{c|c}
		\toprule[1.2pt]
		Method & Time(s) \\ 
		\midrule
		SVM & 436.20 \\
		CNN-2D & 687.63 \\ 
		RNN & 586.34 \\ 
		Cross & 634.52 \\ 
		CALC & 792.64 \\ 
		ViT & \textgreater3000 \\ 
		MFT & \textgreater3000 \\ 
		FedFusion(Local)&3426.77\\
		FedFusion & 520.19 \\ 
		\bottomrule[1.2pt]
	\end{tabular}
\end{table}

{\color{black}\subsection{Efficiency Analysis}}

{\color{black}In order to comprehensively compare the efficiency of FedFusion and other comparison methods, we conducted two aspects of validation. Firstly, for the toy experiment, we measured the necessity of SVD module in three off-site HSI/LiDAR datasets. The experimental results show that before using SVD, the communication cost on the Houston 2013 dataset, Trento dataset, and MUUFL dataset was 45.97MB, 59.58MB, and 110.26MB, respectively. However, the communication cost of these datasets is 14.68MB, 14.75MB, and 26.21MB, respectively. Communication cost is reduced by more than three times, greatly improving FedFusion's processing efficiency.
	Secondly, to verify the efficiency of FedFusion on the simulated satellite system, we first set up a federated satellite platform with the same hardware setup as the real satellite system and the live satellite data. We then compare the training and inference methods under the same computational resources and verify the efficiency superiority of FedFusion by comparing the time of the different methods. As shown in Table \ref{speed}, compared to the local single-client running, FedFusion takes full advantage of multi-client federated learning, reducing training time from 3426.77 seconds to 520.19 seconds, achieving a 6.59$\times$ acceleration. Compared to other deep learning methods, FedFusion has a lower running time than classical networks such as CNN-2D, RNN, and Cross. However, Transformer models (e.g. ViT, MFT) are difficult to deploy in satellite onboard resources due to their huge computational complexity, which takes more than 3000 seconds. The above experiments all verify the high efficiency of FedFusion compared with other methods.}

\subsection{Visual Comparison}

The results in Figs. \ref{vis-h}, \ref{vis-t} and \ref{vis-m} demonstrate our macroscopic performance evaluation on the classification maps generated by various models. To achieve this, we utilized a visualization method that assigned a unique color to each class. Traditional classifiers produce distinct classification results. However, their multimodal fusion methods only rely  on the spectral representation of HSI data, which lack spatial information and result in prominent salt and pepper noise in the classification maps. In comparison, the transformer method produces visually superior results in the classification maps when compared to classic networks. The transformer's network architecture prioritizes the correlation between adjacent spectra and channels, and it can transmit location information across layers, contributing to its enhanced performance. On the other hand, FedFusion approaches the problem by considering the complementary information from different modal sources through its feature-level fusion characteristic. This approach reduces the granularity of texture in the classification maps, resulting in a more diversified and refined set of details. Overall, FedFusion is suitable for generating classification maps with improved performance and nuanced details. Its asynchronous federated optimization algorithm and feature-level fusion approach make it the ideal choice for land use and scene classification applications.


\section{Conclusion}

{\color{black}A key issue in the federated feature fusion problem is how to develop a unified model for feature fusion and global update without losing low-rank features of high-dimensional data, which has not been adequately solved under the previous framework. This paper presents the FedFusion framework, which effectively addresses the challenges of multi-modality fusion in multi-satellite and multi-satellite constellations for remote sensing telemetry tasks.  FedFusion utilizes manifold learning and federated learning techniques, combined with a dynamic feature approximation method by singular value decomposition, to enable in-orbit fusion of multi-satellite and multi-modality without the need to transmit back to the ground.  The proposed method achieves excellent performance on multimodal data and exhibits significant advantages for generalized multi-modality learning problems.  

Moreover, accomplishing multimodal data in-orbit fusion is a pivotal task in remote sensing intelligent interpretation. The proposed FedFusion method successfully performs manifold-driven feature-level fusion on multimodal data. It can also optimize communication and achieve optimal performance across the datasets considered. Hence, it is agreed that FedFusion demonstrates exceptional fusion ability and communication performance, showing its potential for executing federated interpretation tasks in remote sensing and earth observation. It is worth noting that our unit is about to launch a number of hyperspectral and LiDAR satellites, which are equipped with FT-D2000 processors. FedFusion has completed in-orbit pre-verification on 8 FT-D2000 industrial module, and the algorithm is about to be mounted on satellites for real in-orbit communication and processing. We will also continue to update the progress of our in-orbit federal work, and we welcome your continued attention and valuable suggestions for us.}


%
\bibliographystyle{IEEEtran}
\bibliography{reference}
\newpage

%
%
%
%

\vfill

\end{document}